\documentclass[a4paper]{article}
\usepackage[square,sort,comma,numbers]{natbib}
\usepackage{algorithm}
\usepackage[noend]{algpseudocode}
\usepackage[english]{babel}
\usepackage[utf8x]{inputenc}
\usepackage[T1]{fontenc}
\usepackage{mathrsfs}
\usepackage{amsfonts}
\usepackage{float}
\usepackage{mwe}
\usepackage{natbib}
\usepackage{graphicx}
\usepackage{subfig}
\usepackage{mwe}
\usepackage{bbm}
\usepackage[a4paper,top=2cm,bottom=2cm,left=2cm,right=2cm,marginparwidth=1.5cm]{geometry}

\usepackage{amsmath}
\usepackage{graphicx}
\usepackage[colorinlistoftodos]{todonotes}
\usepackage[colorlinks=true, allcolors=blue]{hyperref}
\usepackage{color}

\usepackage{amsthm}
 
 \makeatletter
\def\BState{\State\hskip-\ALG@thistlm}
\makeatother
\theoremstyle{definition}

\title{A Distribution Similarity Based Regularizer for Learning Bayesian Networks \\
\large{Its Application on Modeling Wave Propagation in Inhomogeneous Media}}

\begin{document}
\author{Weirui Kong \href{mailto:weiruik@cs.ubc.ca}{weiruik@cs.ubc.ca} \quad Wenyi Wang \href{mailto:wenyiw@cs.ubc.ca}{wenyiw@cs.ubc.ca} \\
Department of Computer Science, University of British Columbia}\maketitle
\begin{abstract}
 Probabilistic graphical models compactly represent joint distributions by decomposing them into factors over subsets of random variables. In Bayesian networks, the factors are conditional probability distributions. For many problems, common information exists among those factors. Adding similarity restrictions can be viewed as imposing prior knowledge for model regularization. With proper restrictions, learned models usually generalize better. In this work, we study methods that exploit such high-level similarities to regularize the learning process and apply them to the task of modeling the wave propagation in inhomogeneous media. We propose a novel distribution-based penalization approach that encourages similar conditional probability distribution rather than force the parameters to be similar explicitly. We show in experiment that our proposed algorithm solves the modeling wave propagation problem, which other baseline methods are not able to solve. 
\end{abstract}
\section{Introduction}
    Probabilistic graphical models compactly represent joint distributions by decomposing them into factors over subsets of random variables. In Bayesian networks, the factors are conditional probability distributions. For many problems, common information exists among those factors. For instance, the simplified Ising model (a typical graphical model of ferromagnetism) restricts the parameters of all local potentials and all of the neighbor interactions to be identical \cite{mccoy2014two}. Adding similarity restrictions can be viewed as imposing prior knowledge for model regularization. With proper restrictions, learned models usually generalize better. 
    
However, for problems with inhomogeneous space, the identical assumption oversimplifies the problem. But still, we believe that common information would exist in a higher level. In this work, we study methods that exploit such high-level similarities to regularize the learning process and apply them to the task of modeling the wave propagation in inhomogeneous media. Mathematically, wave propagation is modeled using differential equations of a perturbation function, and closed form solution does not exist for inhomogeneous material. Using numerical finite difference methods, we generate the dataset that stores the sequential states of the wave propagation in a 2-D grid. This process can be considered as a stationary Markov chain. The task is to learn a transition dynamics that maps the state of current perturbations to perturbations of next time step. %predict the state of each node at any time $t$ given the state at initial time $t_0$.
    
    We propose a novel distribution-based penalization approach that encourages similar conditional probability distribution rather than force the parameters to be similar explicitly. We also implement three other methods: the "free" model (will be illustrated later), strict parameter sharing and multi-task learning, and show that our approach outperforms these models on our wave propagation dataset.

\section{Related Work}
In this section, we introduce related works of this project. The first two subsections state parameter sharing and multi-task learning as baseline methods. The third subsection provides a brief overview of distribution similarity measures and the formulas as building bocks of proposed regularizer. Lastly we describe the practical problem that we apply and compare the algorithms. 
\subsection{Parameter Sharing}
Architectures based on deep artificial neural networks have improved the state of the art across a wide range of diverse tasks. Most prominently Convolutional Neural Networks have raised the bar on image classification tasks \cite{krizhevsky2012imagenet, simonyan2014very, he2016deep}. Parameter sharing, namely sharing of weights by all neurons in a particular feature map, is crucial to deep CNN models since it controls the capacity of the model and encourages spatial invariance. However, explicitly tying the parameters together may oversimply the problem in some scenarios. Take Ising models as example, the energy of a configuration $\sigma$ is given by the Hamiltonian function $H(\sigma)=-\sum_{i,j}J_{ij}\sigma_i\sigma_j-\mu \sum_{j}h_j\sigma_j$. If the external magnetic field is homogeneous, it’s reasonable to simplify the model by setting $h_j=h$ for all $j$ and $J_{ij}=J$ for all $i,j$ paris. When the external magnetic field is inhomogeneous, whereas, strict parameter sharing does not refer to reality anymore. A model with strict parameter sharing is likewise too naïve to approximate the wave propagation problem, nevertheless we implement such model as a baseline.
\subsection{Multi-task Learning}
Multi-Task Learning (MTL) is a learning paradigm in machine learning and its target is to leverage useful information contained in multiple related tasks to help improve the generalization performance of all the tasks. Consider the wave propagation problem, we can formulize it as finding the distribution $P(x_t|x_{t-1})$, where $x_t$ denotes the state of all nodes at time $t$. The assumption is that the state of each node at time $t$ is independent of other nodes at time $t-1$ given the states of its neighbors at time $t-1$. So the distribution can be decomposed into $N$ (assume we have $N$ nodes in total) conditional distributions. Rather than using $N$ neural networks to approximate those distributions, MTL uses one neural network with $N$ outputs. The first few hidden layers are shared among nodes, while the following layers are node-specific. \cite{baxter1997bayesian} shows that such setting greatly reduces the risk of overfitting. This makes sense intuitively: The more factors we are learning simultaneously, the more our model has to find a representation that captures all of the factors and the less is the chance of overfitting. We implement a MTL model as the opponent of our distribution-based penalization model.
\subsection{Similarity Measure Between Distributions}
Statistical distance measures dissimilarity between two probabilistic distributions. It was originally developed for testing samples follow hypothesized distributions \cite{thas2010comparing}. In our work, the targets are to learn conditional distributions, and penalize on their dissimilarities. \cite{martos2015statistical} provides a comprehensive overview of this field. Typical statistical distances can be classified as metric and non-metric measures. For our purpose, metric measures are easier to work with since many optimization algorithms are developed in metric space, and symmetry simplifies the possibilities. However the non-metric measures often encode desired information such as KL divergence measures information gain. In this work, we employ KL divergence and Bhattacharyya distance for regularization. Since we work on Gaussian random fields, the difference measures exist in closed forms. For fixed $\mu_1, \sigma_1, \mu_2, \sigma_2$, which are the means and standard deviations that define two normal distributions, the KL divergence and Bhattacharyya distance are
\begin{align*}
KL(\mu_1, \sigma_1, \mu_2, \sigma_2) &= \ln(\sigma_2) - \ln(\sigma_1) + \frac{\sigma_1^2+(\mu_1-\mu_2)^2}{2\sigma_2^2} - \frac{1}{2}, \text{and}\\
%log(s2) - log(s1) + (s1**2+(mu1-mu2)**2)*.5/s2**2 - .5
BH(\mu_1, \sigma_1, \mu_2, \sigma_2) &= \frac{1}{4} \ln(\frac{1}{4}(\frac{\sigma_1^2}{\sigma_2^2} + \frac{\sigma2^2}{\sigma_1^2} + 2)) + \frac{(\mu_1-\mu_2)^2}{\sigma_1^2+\sigma_2^2}.
%.25*(log(.25*(s1**2/s2**2+s2**2/s1**2+2)) + (mu1-mu2)**2/(s1**2+s2**2))
\end{align*}
Notice that Bhattacharyya distance is symmetric.

\subsection{Wave Propagation in Inhomogeneous Media and Damage Identification}
The problem of damage identification using ultrasonic waves is to find singular material properties location by analyzing ultrasonic waves propagating through the material. This problem can be applied to monitoring health of structure materials such as aircrafts. The difficulty of this problem is that the material properties as a function over space interacts with the wave through a partial differential equation. A closed form solution of the equation does not exist. Therefore, learning the material properties from observations of the wave function is very hard. 

For two-dimensional elastic wave propagation problem, P-SV wave and its approximation using finite difference method was proposed in \cite{virieux1986p}. Later work shows finite difference method for lamb wave, which is a special case of P-SV wave \cite{gopalakrishnan2016wave}, is useful for the task of damage identification \cite{lee2003modellingA, lee2003modellingB, yan2013bayesian}. Because of the accessibility of source code, we employ a finite difference implementation of P-SV wave \cite{bohlen2016three} for data generation.

    Existing methods for damage identification are limited in the following perspectives. First, many of existing works are not quantitative. For example \cite{lee2003modellingB} systematically studies the wave behavior for different configuration of damages, and suggest further measurements and possible damage locations given observed wave behavior. The problems of this kind of approaches are that the behaviors are described subjectively, the inferences are inductive, and the results are hard to test. Second, the existing quantitative works rely too much on feature engineering based on simplified models. For example \cite{yan2013bayesian} builds a Bayesian model of damage location and time-of-fly, which is the trivial time of reflective field. To compute the time-of-fly, it is necessary to assume single point damage with strong reflectivity and an infinite homogeneous media. In practice, many of these assumptions oversimplify the problem, and measurements are far from assumed model \cite{wang2015bayesian}. In this work, we propose to learn the local interactions directly from data using graphical models. This will overcome the oversimplification, and a successfully learned graphical model is capable for many kinds of quantitative inference including damage identification.

\section{Problem Specification and Algorithms}
The task we work on is to model the wave propagation on a $d \times d$ grid ($d$ was set to 50 in our experiments). At time $t$, a $50 \times 50$ tensor represents the state (its position) of these 2500 nodes. The problem can be formulized as finding the distribution of the state of all nodes at time $t$, given the state at time $t-1$. We assume the sequential data is a homogeneous Markov chain. The independence assumption is that the state of each node at time $t$ is independent of other nodes at time $t-1$ given the states of its neighbors at time $t-1$. Furthermore, we assume that the state of each node given its neighbors of previous time follows a Gaussian distribution. We use two neural networks to approximate the distribution, one for producing the mean and the other for variance. For a d by d grid, a model without any regularization would contain $d\times d\times2 = 5000$ neural nets. We call it "free" model. The inputs to each neural net are the states of neighbors at time $t-1$, and the output is the mean or variance of the node at time $t$. For all the experiments, we set the neighbors of a node be a $5\times5$ blanket whose center is the chosen node. And because of the conditional independencies, the losses can be easily written as a sum of sub-functions, where each of the sub-functions involves very few of the networks. Therefore, one can use coordinate gradient descent to train those neural nets. 
\subsection{Data Generation}
The training and test data are generated as two time sequences of wave function evaluations at $d\times d$ grid points on a plate. We simulate the wave propagation using the finite difference method that solves P-SV wave equations \cite{fda}.
The densities of plate over the grid points, which is used to solve the differential equation, is set as a constant equals $2200 g/cm^3$ except on the $5\times5$ grids centered at $(23,23)$. The $5\times5$ densities centered at $(23,23)$ are set to 
\[2200 g/cm^3\begin{bmatrix}
    0.5000    &0.3316    &0.2816    &0.3316    &0.5000\\
    0.3316    &0.1876    &0.1408    &0.1876    &0.3316\\
    0.2816    &0.1408    &0.0704    &0.1408    &0.2816\\
    0.3316    &0.1876    &0.1408    &0.1876    &0.3316\\
    0.5000    &0.3316    &0.2816    &0.3316   &0.5000\\
\end{bmatrix}.\] The velocities on the grid points are set to $3000 m/s$. The training and test data are simulated on this material with different wave source points for $1.5$ seconds. The data of first $0.5s$ are truncated for each of the sequences. The training data at time $0.6s$ is visualized in figure 1.

\begin{figure}[h]
\centering
\includegraphics[width=.8\textwidth]{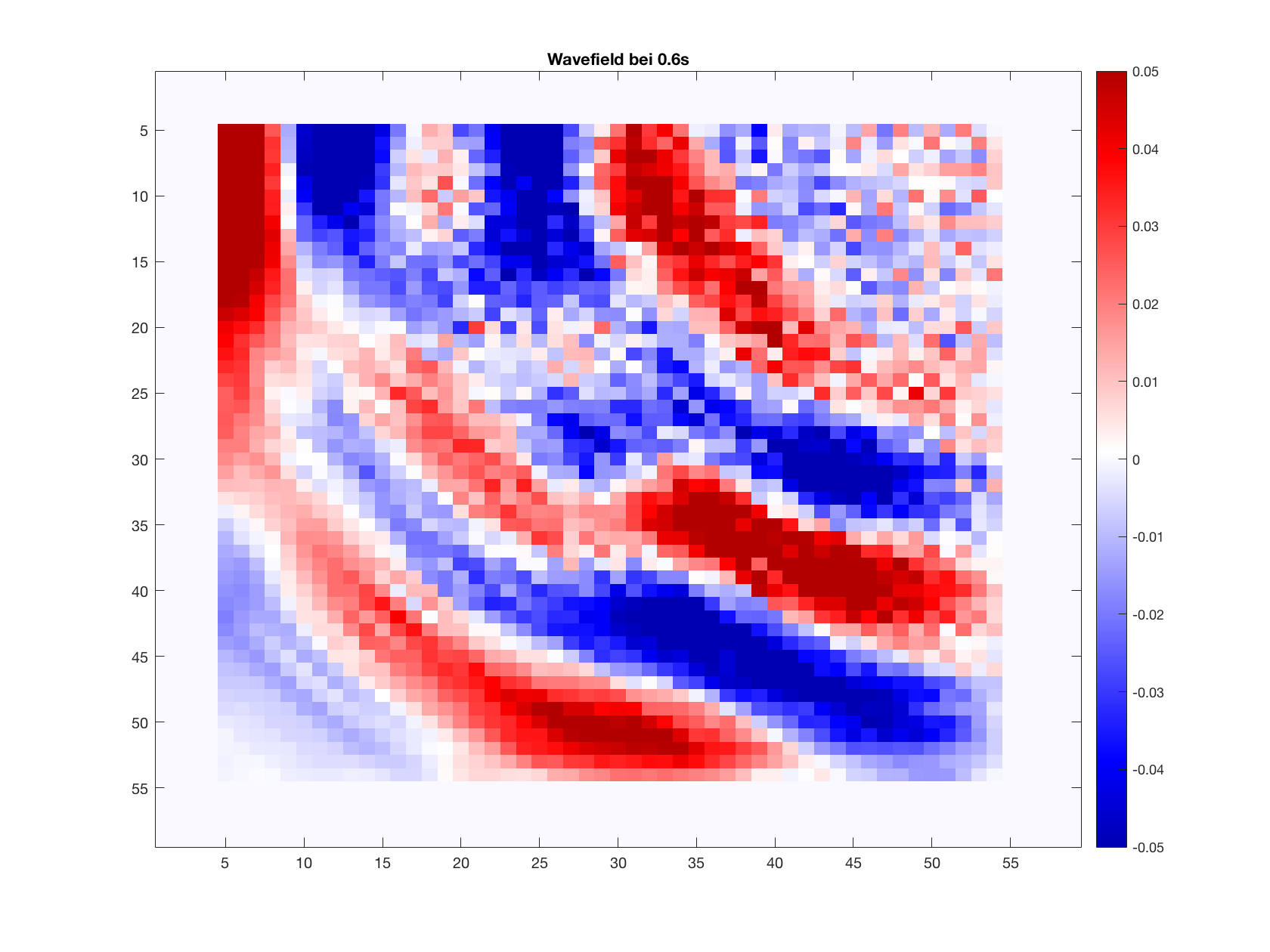}
\caption{The visualization of training data at time $0.6$s.}
\end{figure}

\subsection{Baseline Algorithms}
We implement three baseline methods, i.e., free model, model with strict parameter sharing and multi-task learning model. The free model consists of $50\times 50\times 2=5000$ neural nets, computing the mean and variance for the distribution of each node individually. Figure 2.a shows the structure of one such neural net. For the  model with strict parameter sharing, we use two neural nets to approximate the distribution of conditional mean and variance for all the nodes. It can be regarded as tying the weights of 5000 neural nets all together. The assumption is strong: given the state of its neighbors at time $t-1$, the distribution of the state of each node is identical, regardless the location of node. In this setting, the model is trained on more data (For one sequential data, we only need to train 2 neural nets rather than $d\times d\times 2$). It's easier to converge and less likely to overfit. But the negative point of this model is that it oversimplifies the problem by ignoring the inhomogeneous property of the material. In another word, it makes a homogeneous assumption. Our last baseline model uses multi-task learning technique. We have one neural net for computing the mean of all the nodes and one single neural net for the variance. The first hidden layer is shared but the second hidden layer corresponds to different distributions of each node. Figure 2.b shows the structure of MLT based model. This model exploits the common information among the grid nodes to some extend without simplifying the problem too much.

\begin{figure}[h]
\centering
\subfloat[Single neural net structure]
{
	\includegraphics[width=.4\linewidth]{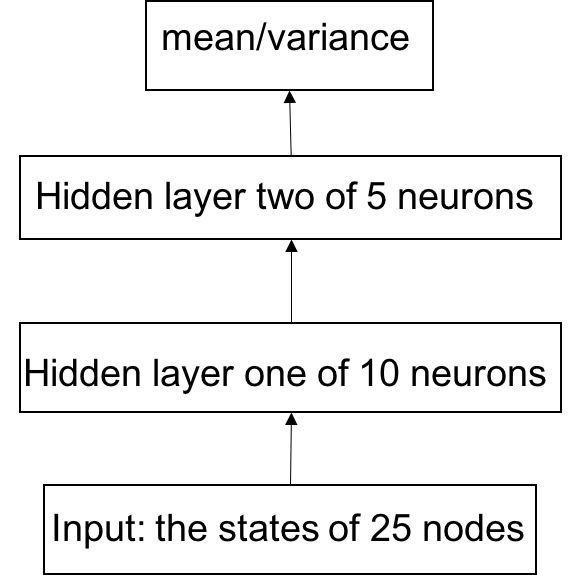}
	\label{fig:1d_standard}
}
\subfloat[Multi-task learning structure]
{
	\includegraphics[width=.6\linewidth]{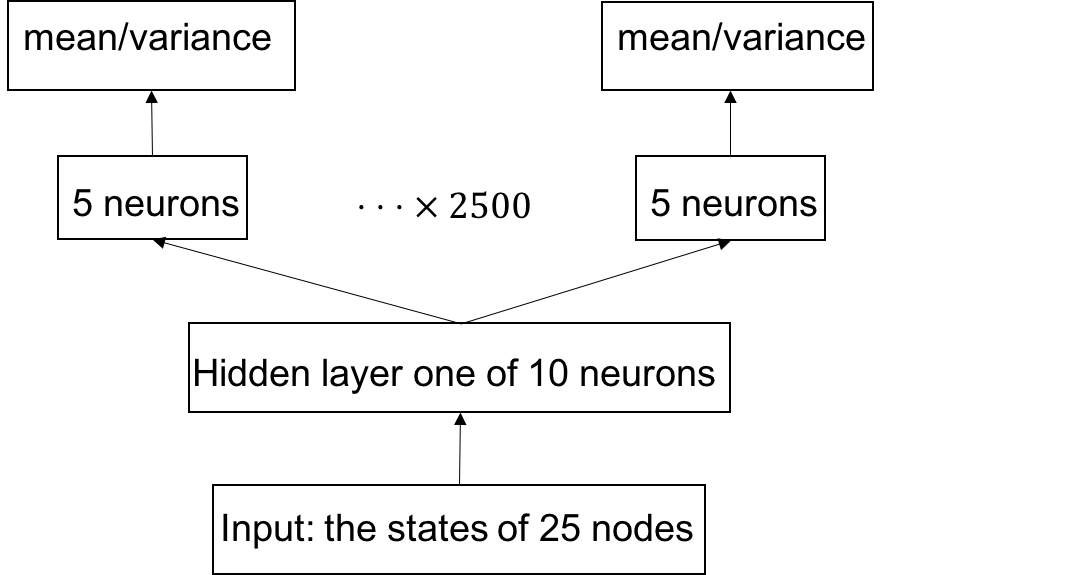}
	\label{fig:1d_monot}
}
\caption{The neural net architectures}
\label{fig:1d}
\end{figure}

\subsection{Distribution Similarity Based Penalization}
Our proposed model is the same as the aforementioned free model, except that we add distribution-based penalization. Now we still have $d\times d\times 2$ neural nets, but we restrict the neural nets that approximate the distributions of nearby nodes to be similar. The regularization is at the distribution level, that is, the statistical distance between two conditional Gaussians. More specifically, there are two neural nets being associated with every node, whose outputs are the mean and variance respectively. The loss function consists of negative log likelihood and a penalization term. The penalization is the closed form statistical distances (a function of two pairs of mean and variance) among nearby Gaussians conditioned on some identical neighborhood values. The neighborhood values are selected in different ways.
%This will be explained in the experimental results section. 
Mathematically, the loss function is written as
\begin{align}
loss(\Theta) = -L(D|\Theta) + \lambda \sum_t \sum_i\sum_{j \in NB(i)} P(\mu(X_{i,j,t}|\Theta_i), \sigma(X_{i,j,t}|\Theta_i), \mu(X_{i,j,t}|\Theta_j), \sigma(X_{i,j,t}|\Theta_j))
\end{align} 
where $\Theta$ are the parameters that define the neural nets, $L$ is the log likelihood function and $P$ is a penalization function that measures the difference between two normal distributions given the parameters. The hyperparamter $\lambda$ is set to one for all the experiments. Notice that for each node the two networks define a conditional normal distribution, and the distribution difference measures only apply on unconditioned probability distributions. The conditional distributions' differences need to be measured given the parent data. In the loss function, at each time step $t$, we regularize the parameters by summing the distribution difference measures given parent values $X_{i,j,t}$  over all neighborhoods $i,j$. The parent values (i.e. the $X_{i,j,t}$'s) can be selected in different ways. One is to randomly select $5 \times 5$ neighborhood values over the whole dataset. The other way is to select the parent values as the $5\times5$ neighborhood values at time $t-1$ whose center is node $i$. We employ two statistical distances for experiments: the KL divergence and Bhattacharyya distance. 
\section{Experimental Results}
We use an epoch of 3, a learning rate of 0.01 and Adam optimizer to train all models. Both the training data and test data contain $50\times 50$ node states of 119 time steps. For each model, we record the training loss (negative log likelihood) and test loss at every time step. Since these values are in a great magnitude, we process them by taking $\ln$ to the values. For negative values $x$, $-\ln(-x)$ are used. For $|x|<1$, we simply display a zero. We create figures with an x-axis denoting time steps (119 per epoch $\times$ 3 epochs) and y-axis denoting the log value of loss. 

    Figure 3.a and 3.b show the training and test loss of the free model. One can barely tell if the model converges or not from the training loss for it just fluctuates violently. From the test loss we can see that the overall performance is unstable: it achieves a low loss in the second epoch but the model seems to overfit in the third epoch. 
    
    Figure 4.a and 4.b show the losses of the model with strict parameter sharing. The test loss is large and does not tend to drop as epoch increasing. The reason for this behavior of the loss is this model oversimplifies the problem. 
    
    Figure 5.a and 5.b show the losses of MTL based model. There's no sign of convergence and its losses are larger than the free model. It constrains the model complexity by sharing the first hidden layer, but the second layer alone may not be complex enough to fit the data.
    
    Figure 6.a and 6.b show the losses of KL divergence regularized model. When computing equation (1), we pick parent values $X_{i,j,t}$ as the $5\times 5$ neighborhood values of previous time centered at node $i$. We can see that the regularized model converges much faster. It also effectively avoids the risk of overfitting since the test loss keeps decreasing as the model is trained with more epochs.
    
    Figure 7.a and 7.b show the losses of Bhattacharyya distance regularized model. Again, the regularized model boosts the convergence and makes the training more stable. For the test loss, however, we do not see the similar pattern of Figure 6.b. So we turn to another strategy of picking the parent values $X_{i,j,t}$ by randomly selecting $5 \times 5$ neighborhood values over the whole dataset. Figure 8 shows the results in this setting. Now the Bhattacharyya distance regularized model behaves similar to the KL divergence one, though the latter is more stable. A possible interpretation of this observation is that since the purpose of evaluating penalization given $X_{i,j,t}$ is to make the two conditional probabilities similar for all possible parent values, using the same $X_{i,j,t}$ as the input to the neural nets makes it biased. 
    
\begin{figure}[H]
\centering
\subfloat[Training loss]
{
	\includegraphics[width=.49\linewidth]{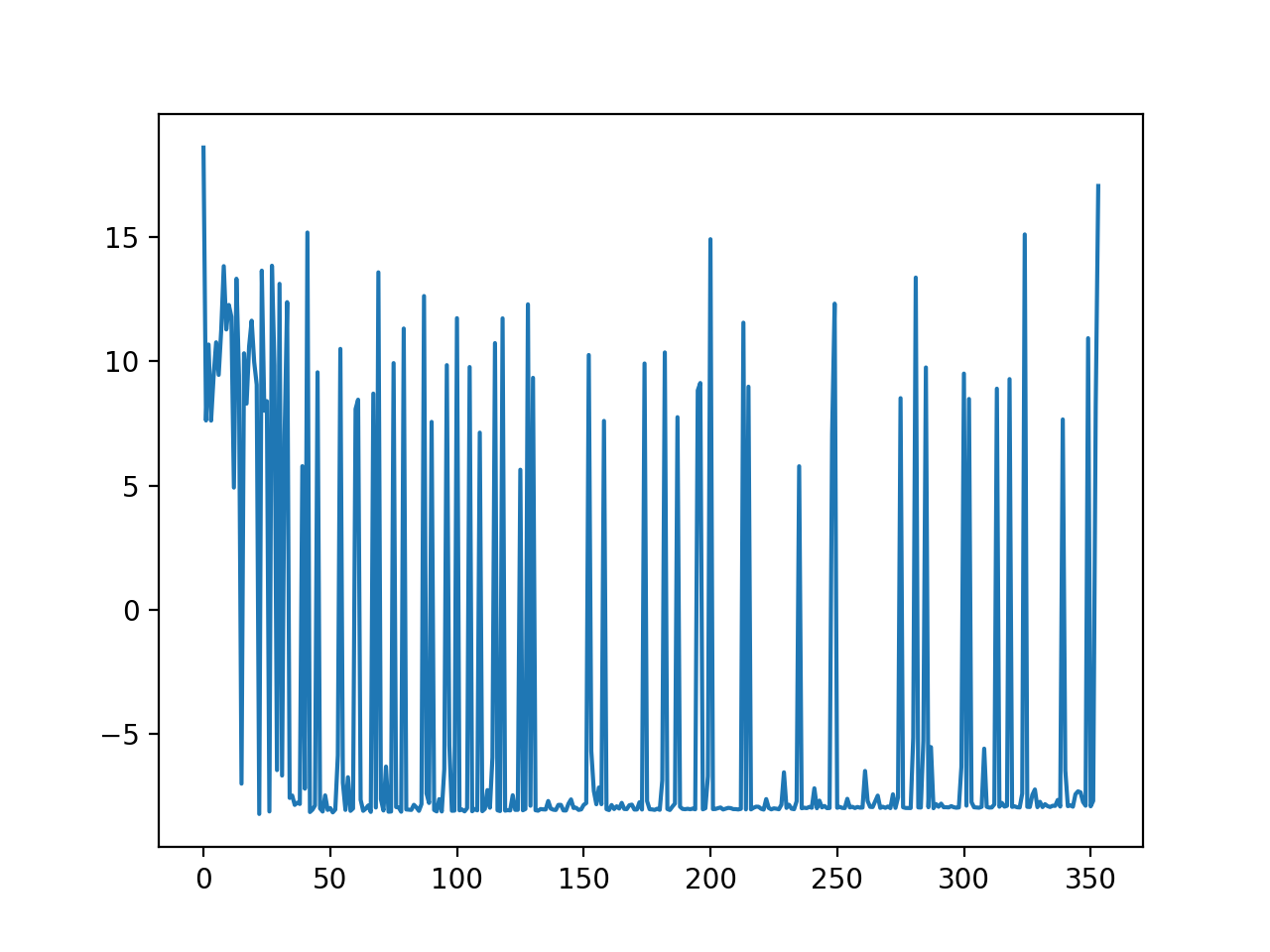}
	\label{fig:1d_standard}
}
\subfloat[Test loss]
{
	\includegraphics[width=.49\linewidth]{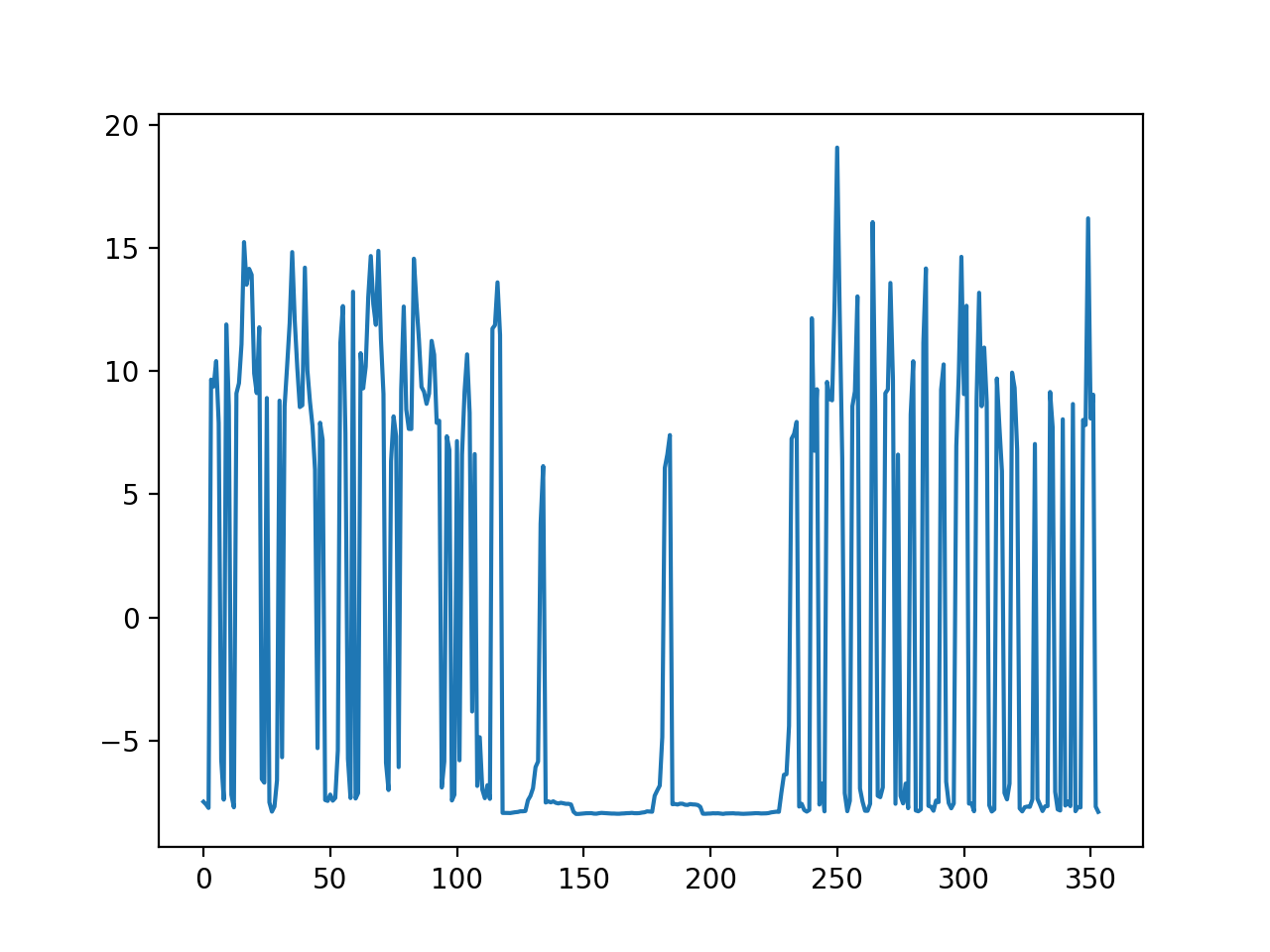}
	\label{fig:1d_monot}
}
\caption{The free model}
\label{fig:1d}
\end{figure}

\begin{figure}[H]
\centering
\subfloat[Training loss]
{
	\includegraphics[width=.49\linewidth]{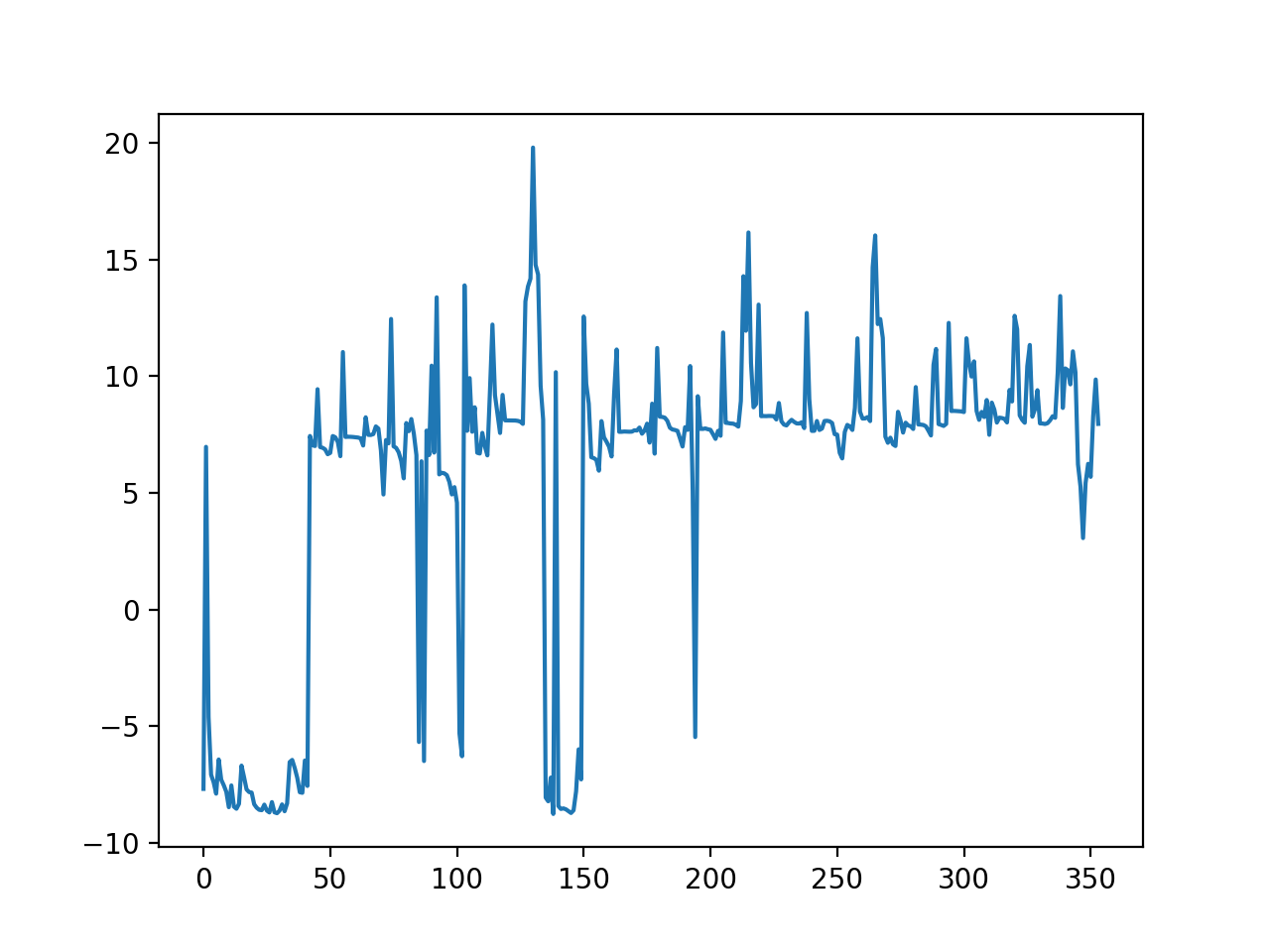}
	\label{fig:1d_standard}
}
\subfloat[Test loss]
{
	\includegraphics[width=.49\linewidth]{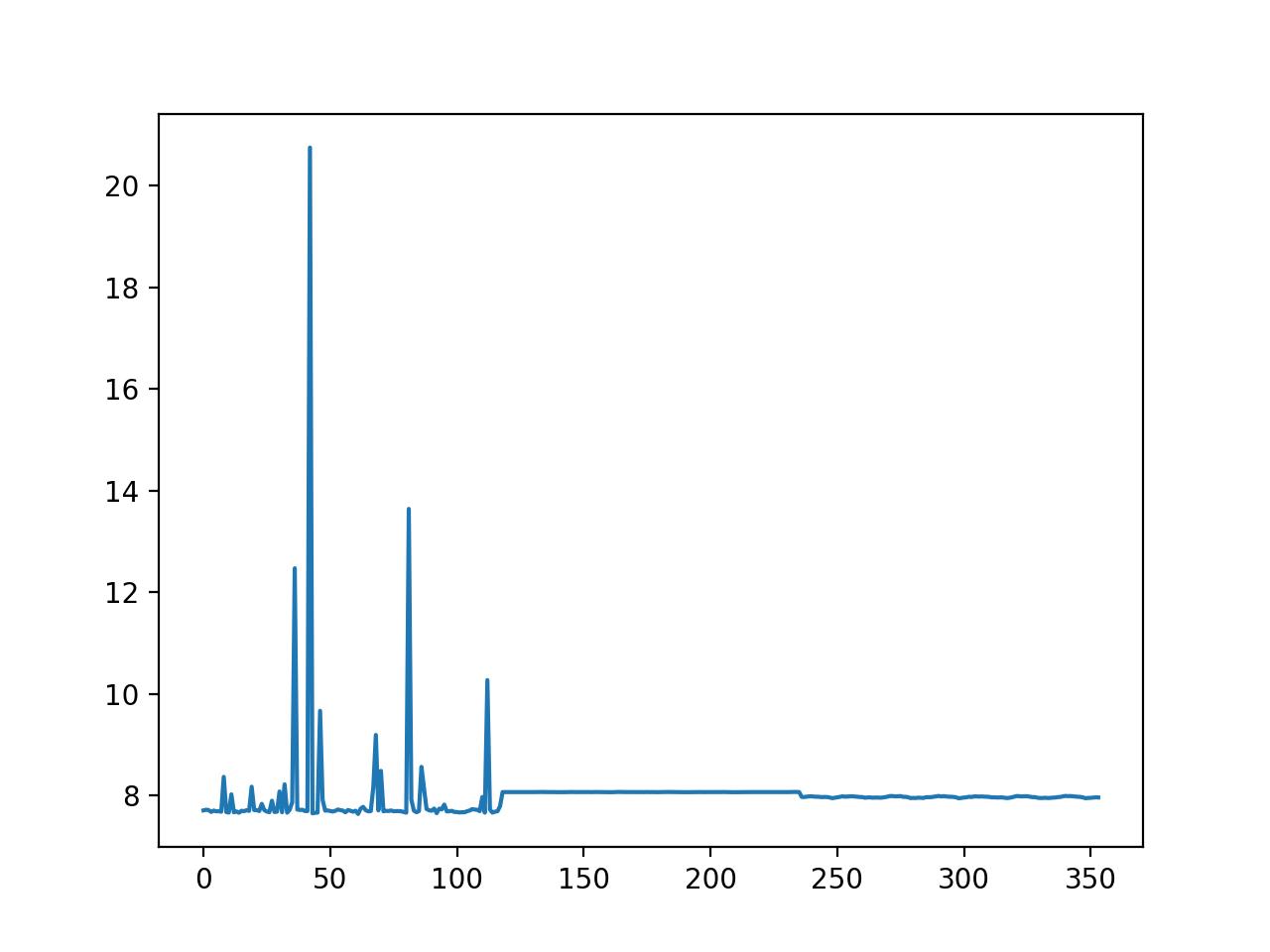}
	\label{fig:1d_monot}
}
\caption{Model with strict parameter sharing}
\label{fig:1d}
\end{figure}

\begin{figure}[H]
\centering
\subfloat[Training loss]
{
	\includegraphics[width=.49\linewidth]{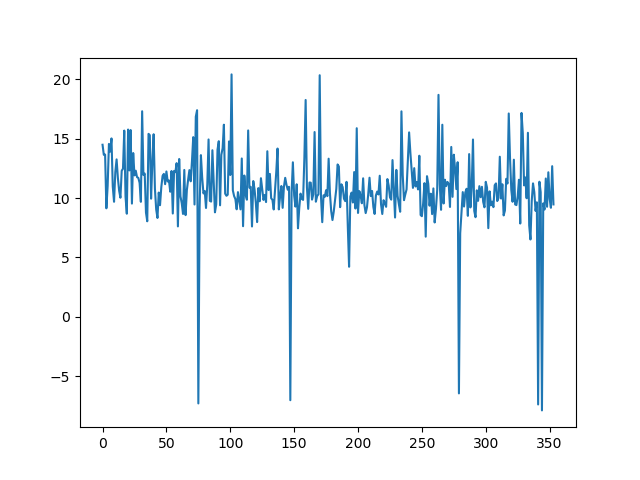}
	\label{fig:1d_standard}
}
\subfloat[Test loss]
{
	\includegraphics[width=.49\linewidth]{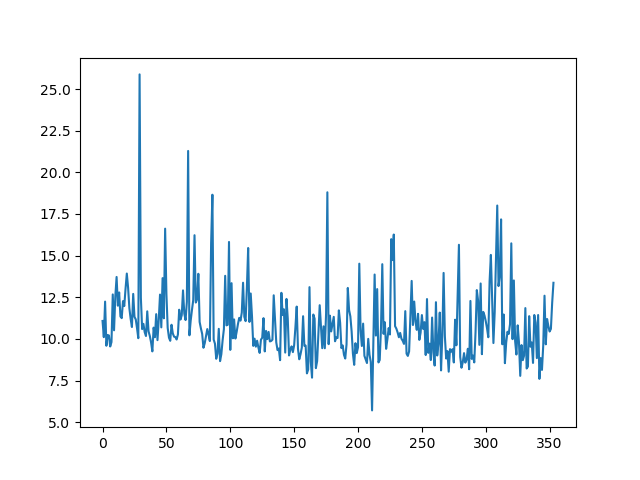}
	\label{fig:1d_monot}
}
\caption{The MTL model}
\label{fig:1d}
\end{figure}

\begin{figure}[H]
\centering
\subfloat[Training loss without regularization term]
{
	\includegraphics[width=.49\linewidth]{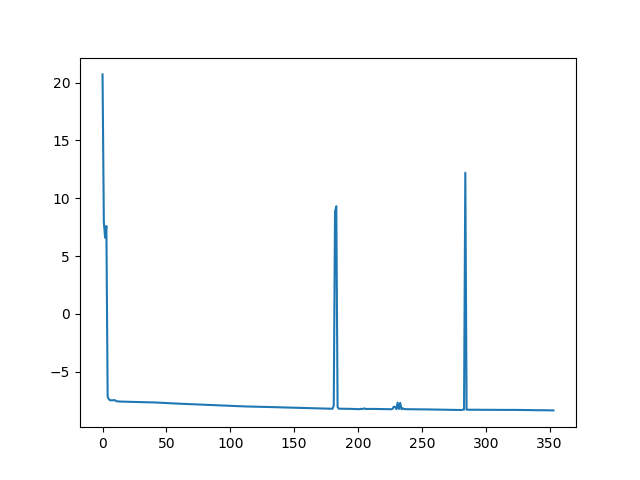}
	\label{fig:1d_standard}
}
\subfloat[Test loss]
{
	\includegraphics[width=.49\linewidth]{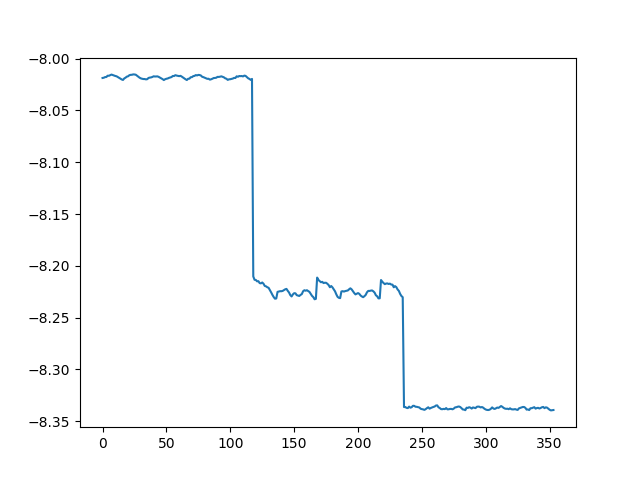}
	\label{fig:1d_monot}
}
\caption{KL divergence regularized model}
\label{fig:1d}
\end{figure}

\begin{figure}[H]
\centering
\subfloat[Training loss without regularization term]
{
	\includegraphics[width=.49\linewidth]{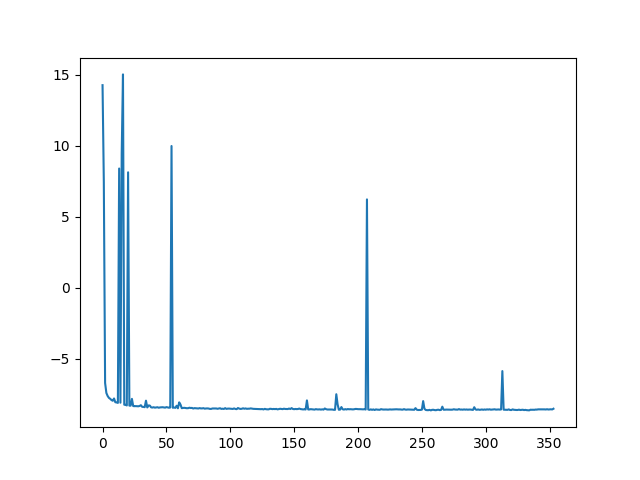}
	\label{fig:1d_standard}
}
\subfloat[Test loss]
{
	\includegraphics[width=.49\linewidth]{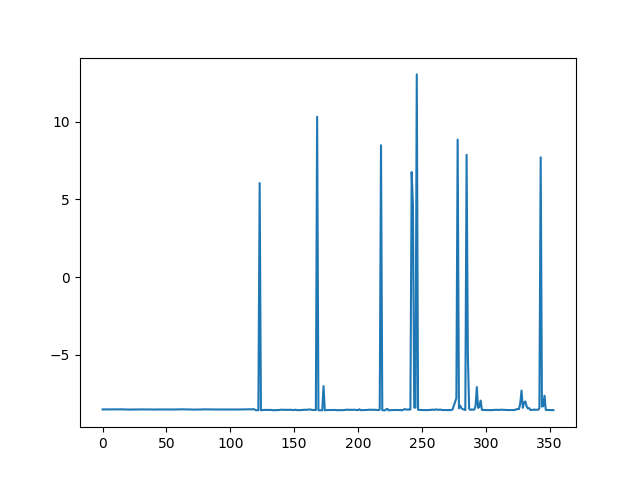}
	\label{fig:1d_monot}
}
\caption{Bhattacharyya distance regularized model}
\label{fig:1d}
\end{figure}

\begin{figure}[H]
\centering
\subfloat[Training loss without regularization term]
{
	\includegraphics[width=.49\linewidth]{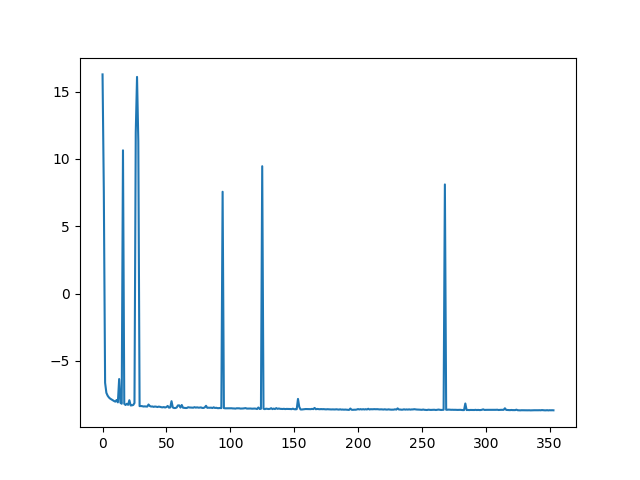}
	\label{fig:1d_standard}
}
\subfloat[Test loss]
{
	\includegraphics[width=.49\linewidth]{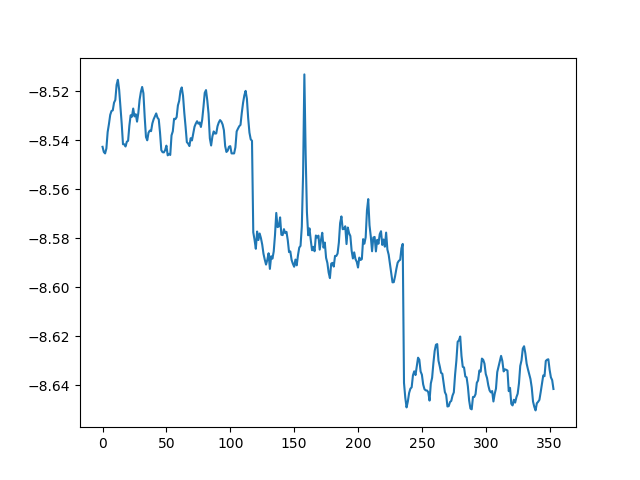}
	\label{fig:1d_monot}
}
\caption{Bhattacharyya distance regularized model (randomly picked $X_{i,j,t}$)}
\label{fig:1d}
\end{figure}

\section{Conclusion}
Modeling wave propagation requires learning a numerous number of conditional probabilities from limited data. The experiments show that the three baseline algorithms do not solve this problem. The reasons for the "free" and strict parameter sharing model do not perform well is trivial. For the multi-task learning method using layer-sharing neural nets, comparing to our method, it dose not use the spacial information to reduce the model complexity. It uses a single common layer shared by all nodes instead of various common layers shared by neighbors. Because of the difficulty of implementation, we did not implement a parameter similarity based penalization. It will be more fair to compare our proposed distribution similarity based penalization with a parameter based one since they could use the same amount of prior knowledge of the problem, i.e. the spacial information. However the results we provide is sufficient to show the proposed penalization method is practically valuable. 

We test our algorithm with wave function observation data on all grid points to avoid the difficulty of learning with latent variables. In practice, it is more likely that there are only tens or hundreds of the wave function locations (possibly vary over time) available. A algorithm that builds the transition model of the wave with unobserved data will be more significant from a practical point of view. This work does not limit to the damage identification problem. It would generalize to other systems that can be solved by finite difference methods with simple modifications.

In summary, we propose a distribution similarity base penalization function to regularize conditional probability distributions learned for Bayesian networks. The experimental results are done for the problem of modeling a wave function transition, where the model is valuable for material damage identification. With our regularization, the conditional probabilities as neural nets that define a Bayesian net is successfully learned, which are failed using other three baseline methods.
\bibliography{ref}

\newcommand{\etalchar}[1]{$^{#1}$}
\begin{thebibliography}{WJT{\etalchar{+}}15}

\bibitem[Bax97]{baxter1997bayesian}
Jonathan Baxter.
\newblock A bayesian/information theoretic model of learning to learn via
  multiple task sampling.
\newblock {\em Machine learning}, 28(1):7--39, 1997.

\bibitem[BW16]{bohlen2016three}
Thomas Bohlen and Florian Wittkamp.
\newblock Three-dimensional viscoelastic time-domain finite-difference seismic
  modelling using the staggered adams--bashforth time integrator.
\newblock {\em Geophysical Journal International}, 204(3):1781--1788, 2016.

\bibitem[Gop16]{gopalakrishnan2016wave}
Srinivasan Gopalakrishnan.
\newblock {\em Wave propagation in materials and structures}.
\newblock CRC Press, 2016.

\bibitem[HZRS16]{he2016deep}
Kaiming He, Xiangyu Zhang, Shaoqing Ren, and Jian Sun.
\newblock Deep residual learning for image recognition.
\newblock In {\em Proceedings of the IEEE conference on computer vision and
  pattern recognition}, pages 770--778, 2016.

\bibitem[Inc13]{fda}
GitHub Inc.
\newblock Finite-difference seismic wave simulation.
\newblock \url{https://github.com/florianwittkamp/FD_ACOUSTIC}, 2013.

\bibitem[KSH12]{krizhevsky2012imagenet}
Alex Krizhevsky, Ilya Sutskever, and Geoffrey~E Hinton.
\newblock Imagenet classification with deep convolutional neural networks.
\newblock In {\em Advances in neural information processing systems}, pages
  1097--1105, 2012.

\bibitem[LS03a]{lee2003modellingA}
BC~Lee and WJ~Staszewski.
\newblock Modelling of lamb waves for damage detection in metallic structures:
  Part i. wave propagation.
\newblock {\em Smart Materials and Structures}, 12(5):804, 2003.

\bibitem[LS03b]{lee2003modellingB}
BC~Lee and WJ~Staszewski.
\newblock Modelling of lamb waves for damage detection in metallic structures:
  Part ii. wave interactions with damage.
\newblock {\em Smart Materials and Structures}, 12(5):815, 2003.

\bibitem[MV15]{martos2015statistical}
Gabriel~Alejandro Martos~Venturini.
\newblock Statistical distances and probability metrics for multivariate data,
  ensembles and probability distributions.
\newblock 2015.

\bibitem[MW14]{mccoy2014two}
Barry~M McCoy and Tai~Tsun Wu.
\newblock {\em The two-dimensional Ising model}.
\newblock Courier Corporation, 2014.

\bibitem[SZ14]{simonyan2014very}
Karen Simonyan and Andrew Zisserman.
\newblock Very deep convolutional networks for large-scale image recognition.
\newblock {\em arXiv preprint arXiv:1409.1556}, 2014.

\bibitem[Tha10]{thas2010comparing}
Olivier Thas.
\newblock {\em Comparing distributions}.
\newblock Springer, 2010.

\bibitem[Vir86]{virieux1986p}
Jean Virieux.
\newblock P-sv wave propagation in heterogeneous media: Velocity-stress
  finite-difference method.
\newblock {\em Geophysics}, 51(4):889--901, 1986.

\bibitem[WJT{\etalchar{+}}15]{wang2015bayesian}
Wenyi Wang, Anshul Joshi, Nishith Tirpankar, Philip Erickson, Michael Cline,
  Palani Thangaraj, and Thomas~C Henderson.
\newblock Bayesian computational sensor networks: Small-scale structural health
  monitoring.
\newblock {\em Procedia Computer Science}, 51:2603--2612, 2015.

\bibitem[Yan13]{yan2013bayesian}
Gang Yan.
\newblock A bayesian approach for damage localization in plate-like structures
  using lamb waves.
\newblock {\em Smart Materials and Structures}, 22(3):035012, 2013.

\end{thebibliography}
\bibliographystyle{alpha}
\end{document}